\documentclass[]{interact}
\usepackage{epstopdf}
\usepackage[caption=false]{subfig}

\usepackage{graphicx}
\usepackage{bm}
\usepackage{xcolor}
\usepackage{breqn}
\usepackage{url}

\usepackage[numbers,sort&compress]{natbib}
\bibpunct[, ]{[}{]}{,}{n}{,}{,}
\renewcommand\bibfont{\fontsize{10}{12}\selectfont}
\makeatletter
\def\NAT@def@citea{\def\@citea{\NAT@separator}}
\makeatother

\theoremstyle{plain}

\theoremstyle{definition}

\theoremstyle{remark}

\usepackage{fixltx2e}
\usepackage{algpseudocode}

\usepackage{accents}

\usepackage{ulem}

\MakeRobust{\underaccent} 

\makeatother

\newcommand{\uwidevee}[1]{%
  \mathpalette\douwidevee{#1}%
}

\makeatletter
\newcommand{\douwidevee}[2]{%
  \sbox0{$\m@th#1\widehat{\hphantom{#2}}$}%
  \sbox2{$\m@th#1x$}
  \sbox4{$\m@th#1#2$}
  \dimen0=\ht0
  \advance\dimen0 -.8\ht2
  \dimen2=\dp4
  \rlap{%
    \raisebox{\dimexpr\dimen0-\dimen2}{%
      \scalebox{1}[-1]{\box0}%
    }%
  }%
  {#2}%
}
\makeatother

\usepackage{subcaption}

\begin{document}
\title{Quadrupedal Spine Control Strategies: Exploring Correlations Between System Dynamic Responses and Human Perspectives}
\author{
\name{Nicholas Hafner\textsuperscript{a, b}\thanks{CONTACT Nicholas Hafner Email: nick.h@atr.jp}, Chaoran Liu\textsuperscript{a, c}, Carlos Ishi\textsuperscript{a, c} and Hiroshi Ishiguro\textsuperscript{a, b}}
\affil{\textsuperscript{a}Hiroshi Ishiguro Laboratories, Advanced Telecommunication Research Institute International, Kyoto, Japan; \textsuperscript{b}Intelligent Robotics Laboratory, Osaka University, Osaka, Japan; \textsuperscript{c}Guardian Robot Project, RIKEN, Kyoto, Japan}
}

\maketitle
\begin{abstract}
Unlike their biological cousins, the majority of existing quadrupedal robots are constructed with rigid chassis. This results in motion that is either beetle-like or distinctly robotic, lacking the natural fluidity characteristic of mammalian movements. Existing literature on quadrupedal robots with spinal configurations primarily focuses on energy efficiency and does not consider the effects in human-robot interaction scenarios. Our contributions include an initial investigation into various trajectory generation strategies for a quadrupedal robot with a four degree of freedom spine, and an analysis on the effect that such methods have on human perception of gait naturalness compared to a fixed spine baseline. The strategies were evaluated using videos of walking, trotting and turning simulations. Among the four different strategies developed, the optimised time varying and the foot-tracking strategies were perceived to be more natural than the baseline in a randomised trial with 50 participants. Although none of the strategies demonstrated any energy efficiency improvements over the no-spine baseline, some showed greater footfall consistency at higher speeds. Given the greater likeability drawn from the more natural locomotion patterns, this type of robot displays potential for applications in social robot scenarios such as elderly care, where energy efficiency is not a primary concern.
\end{abstract}


\section{Introduction}

Quadrupedal robotics is a highly active field of research, with more and more powerful robots beginning to be put to use in the field due to their dexterity, mobility and suitability to all-terrain usage scenarios. However a major difference between most modern quadrupedal robots (Unitree A1 \cite{A1_2023}, Spot \cite{Spot_2023}, ANYMal \cite{ANYMal_2016} etc.) and their biological cousins is that biological quadrupeds have an active spine which plays a large role in locomotion \cite{hildebrand_1961, hoyt_1981, bertram_2008}.
Distinct sinusoidal patterns are observable in spine vertebrae, particularly in the sagittal plane \cite{benninger_2004} and transversal plane \cite{benninger_2004, aleotti_2008}. However the impact of motion about the roll axis (frontal plane) is unclear, with different studies reporting negligible movement \cite{benninger_2004} and significant movement \cite{wachs_2016} respectively.

An increase in body mobility from an active spine component could lead to greater mobility in turning and jumping, as well as potentially lead to increases in energy efficiency of gaits. Segmented quadruped spines in robots have been shown to have greater stability and longer flight-phases in bounding motion \cite{eckert_2015}. In addition, in environments where robots need to interact with humans, more natural-looking walking gaits may result in greater uptake. In order to explore a new direction from previous research, this study's objectives are to consider the effect that spine movements and control strategies have upon gait naturalness and system dynamics. This may lead to greater uptake of quadrupedal robots in human-robot interaction (HRI) use cases.

To provide a thorough understanding of the control strategies employed in quadrupedal robots with flexible spines, we discuss existing work in three categories: passive spine systems (Section~\ref{sec:passive_spine_robots}), passive-active spine systems (Section~\ref{sec:hybrid_spine_robots}), and fully active spine systems (Section~\ref{sec:active_spine_robots}). In the following sections the spine configurations, trajectory generation strategies, and control systems of previous works are discussed.

\subsection{Passive Spine Quadrupeds: Exploiting System Dynamics}
\label{sec:passive_spine_robots}
In passive systems, control strategies involve exploiting spine mechanical properties rather than using active actuation, leading to energy efficient motion.
Using a single pitch degree of freedom (DoF) leads to increases in stride length due to spring dynamics \cite{phan_2017}, and increases in the range of velocities available for a fixed energy consumption range \cite{cao_2012}. Using active control in the legs alongside the passive spine joint leads to greater centre of mass (COM) stability \cite{zhang_2013}, while the purely passive system is found to be unstable \cite{cao_2012}.
A single roll DoF is found to lead to different passive gaits (trotting instead of pacing), as well as natural gait transitions to bounding when compared with a rigid spine \cite{nakatani_2009}. Likewise, linear spring-based systems can enable passive cyclic bounding gaits achieved primarily through spine oscillations \cite{koutsoukis_2016}.
As passive spine systems inherently contain no control systems or trajectory generation methods, they are unable to respond to changes in the environment or perform active trajectory following.

\subsection{Hybrid Spine Quadrupeds: Passive-Active Control Strategies}
\label{sec:hybrid_spine_robots}
We define hybrid systems as those with passive spine systems, but some way of changing system dynamic response or otherwise controlling the spine actuation. This enables exploitation of different stiffnesses for different gait speeds \cite{sakai_2007, tsujita_2011, lu_2023}, or mechanically coupling spine flexion to leg movements \cite{hyun_2014}. As optimal spine stiffness varies with gait speed, actuators can be used to change passive system dynamics, enhancing the range of velocities in which motion is stable \cite{sakai_2007, tsujita_2011}. Meanwhile, coupling the spine flexion synchronises the spine movement with gait phases, enabling highly efficient (reaching parity with animals) locomotion \cite{hyun_2014}. The coupled approach matches the expected saggital motion in canine bounding gaits, but is unable to reproduce the biphasal motion observed in the walking gait \cite{wachs_2016}, or transversal vertebrae movement \cite{aleotti_2008}. 
Hybrid approaches do not have the freedom of control of fully actuated systems, and thus cannot actively follow controller-generated trajectories. However they have a greater range of stable gait velocities than purely passive systems \cite{eckert_2015}, and less mass than fully actuated systems. This gives them more adaptability than passive systems, but limited fine motion control.

\subsection{Active Spine Quadrupeds: Greater Control Strategy Freedom}
\label{sec:active_spine_robots}
Active spines enable much broader control strategies, such as using the spine for stabilisation or fluidity of motion. These approaches seek to balance efficiency with utility.
One early approach to active spine locomotion considers a robot with servo-based pitch and yaw DoF robots, determined to have sufficient range of motion to imitate rat movements \cite{ishii_2009}. Robot trajectories are determined based upon the Zero Moment Point theory, such that the robot is statically stable \cite{yamaguchi_1993, vukobratovic_2004}.
In Charlie~\cite{Kuehn_2018}, a 6-DoF Stewart platform~\cite{stewart_1966} is used to enable a large range of motion. Spine trajectories are chosen in order to complement leg motion, reducing the leg joints' ranges of motion and velocities, and are controlled with Proportional-Derivative (PD) controllers. Leg trajectories are once again chosen to be statically stable. A lizard-based quadrupedal system involving a 5-DoF spine plans spine trajectories in order to ensure that the robot COM is always within the foot support triangle \cite{chen_2022}. This allows sharper turning, and greater stability in slippery environments. When augmenting turning, the spine can complement leg motion, but reduction in transversal plane oscillation magnitude is required to prevent self collisions at extreme turning arcs \cite{horvat_2017}.
However, statically stable gaits do not have the performance capabilities of many modern dynamically stable gaits \cite{hyun_2014, ding_2021, li_2023, matsumoto_2023}.
Using a sagittal plane, pneumatically actuated, segmented spine outperforms gait speed of a single rotary DoF spine by a factor of 1.9 \cite{matsumoto_2023}, mirroring hybrid system results \cite{eckert_2015}. This system uses a feed-forward controller with manually chosen timings to synchronise spine and leg locomotion.
By utilising model predictive control (MPC, see Section~\ref{sec:mpc}) linearised about the current spine position, high speed running via flexion and platform-like stabilisation tasks may be performed with a 3-DoF spine \cite{li_2023}. In this system, the spine positions follow sinusoidal patterns, or are directly controlled by the user.
Despite complex control systems, and challenging parameter tuning, active systems are able to directly control the motion generated by the spine system. This often comes with a weight trade-off, as greater motion freedom requires additional actuators. The two approaches observed in literature are manually chosen trajectories, and foot-position dependent trajectories.

\subsection{Human Perception of Walking Motions}

Existing research into spine-enabled quadrupeds broadly focuses on energy efficiency, or locomotion policies. While this is very important, the human perception of the spinal motions associated with such robots is not often explored. Human perception is very important in fields such as social robotics, healthcare, or environments where humans may regularly see or interact with non-industrial robots, and motions play a large role in these perceptions \cite{ishi_2019, ishi_2018}. Miura et al.~\cite{miura_2001} explore the effect of various kinematic properties on human perception of humanoid robot gaits, but do not explore quadrupeds. The uncanny valley effect \cite{mori_2012} is well known to apply to both locomotion and shape of human-like and animal-like robots. Therefore it is important to consider the effects of various spine-centric locomotion strategies from a human-oriented perspective.

\subsection{Contribution}

In this paper we present results of a simulated quadrupedal robot with a four DoF actively actuated spine. Existing work shows that multi-segmented spines better reflect animal motion \cite{eckert_2015}, and our robot incorporates two pitch, and two yaw degrees of freedom. We omit any roll DoF due to ambiguity regarding its function in canine locomotion \cite{benninger_2004, wachs_2016}.

While many existing works consider spine trajectory generation and control strategies, as outlined in the sections above, few consider the effect upon human perception of robot `naturalness' and `likeability'. Our work aims to begin the search for energy efficient, natural-looking gaits, to improve the capacity for animal-inspired robots within HRI.

The research focuses solely on canine spine movements, aiming to develop a naturally perceived walking gait that aligns with canine anatomy, as opposed to alternative forms (e.g. reptilian or feline). Canines are chosen due to their popularity as house pets; the majority of study participants have seen and interacted with canines, giving credibility to `naturalness' scores.

The spine trajectory generation methods investigated include: a varying stiffness approach, a foot position-based augmenting approach, and two fixed set-point methods - one optimised, and one derived from real dog walking data. These approaches integrate with an existing MPC framework \cite{ding_2021}.
Finally, we discuss the implications of spine actuation within quadrupedal robots from a HRI perspective, and propose avenues for future research. 

\section{Simulated Robot Design}

\begin{figure}
    \centering
    \includegraphics[width=0.8\textwidth]{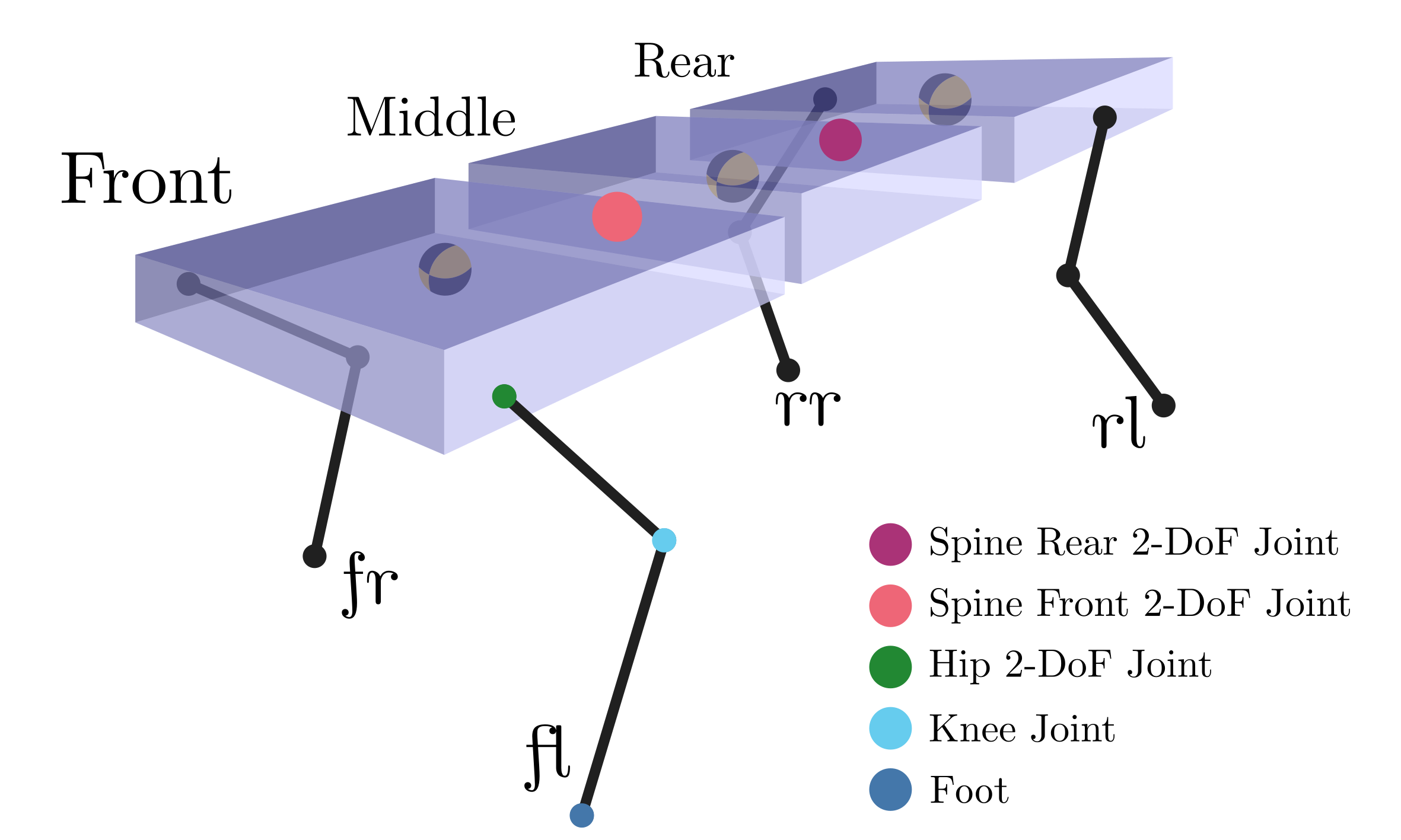}
    \caption{The three rigid bodies and 4 different joints of the simulated spine enabled robot. Note that all legs have identical joint configurations.}
    \label{fig:robot_rigid_body_model}
\end{figure}

In order to rapidly test various spinal configurations, a simulated robot is developed with 2 pitch and 2 yaw degrees of freedom. No roll degree of freedom is considered due to ambiguities about the purpose of roll \cite{benninger_2004, wachs_2016}. We believe that using 4-DoFs strikes a balance between mechanical complexity and actuator weight, and modelling accuracy of the highly-redundant quadrupedal spine. The pitch DoFs can contribute to bounding and climbing motions, while yaw DoFs can reduce turning arc.
The degrees of freedom are aligned such that the robot effectively has two universal joints attached to three large-inertia rigid bodies as shown in a simplified diagram in Figure~\ref{fig:robot_rigid_body_model}. The legs and internal spine joints are modelled with mass and inertia in the simulator, but modelled as mass-less and with negligible inertia in the control system described in Section~\ref{sec:control}.

\begin{figure}
    \centering
    \includegraphics[width=1.0\textwidth]{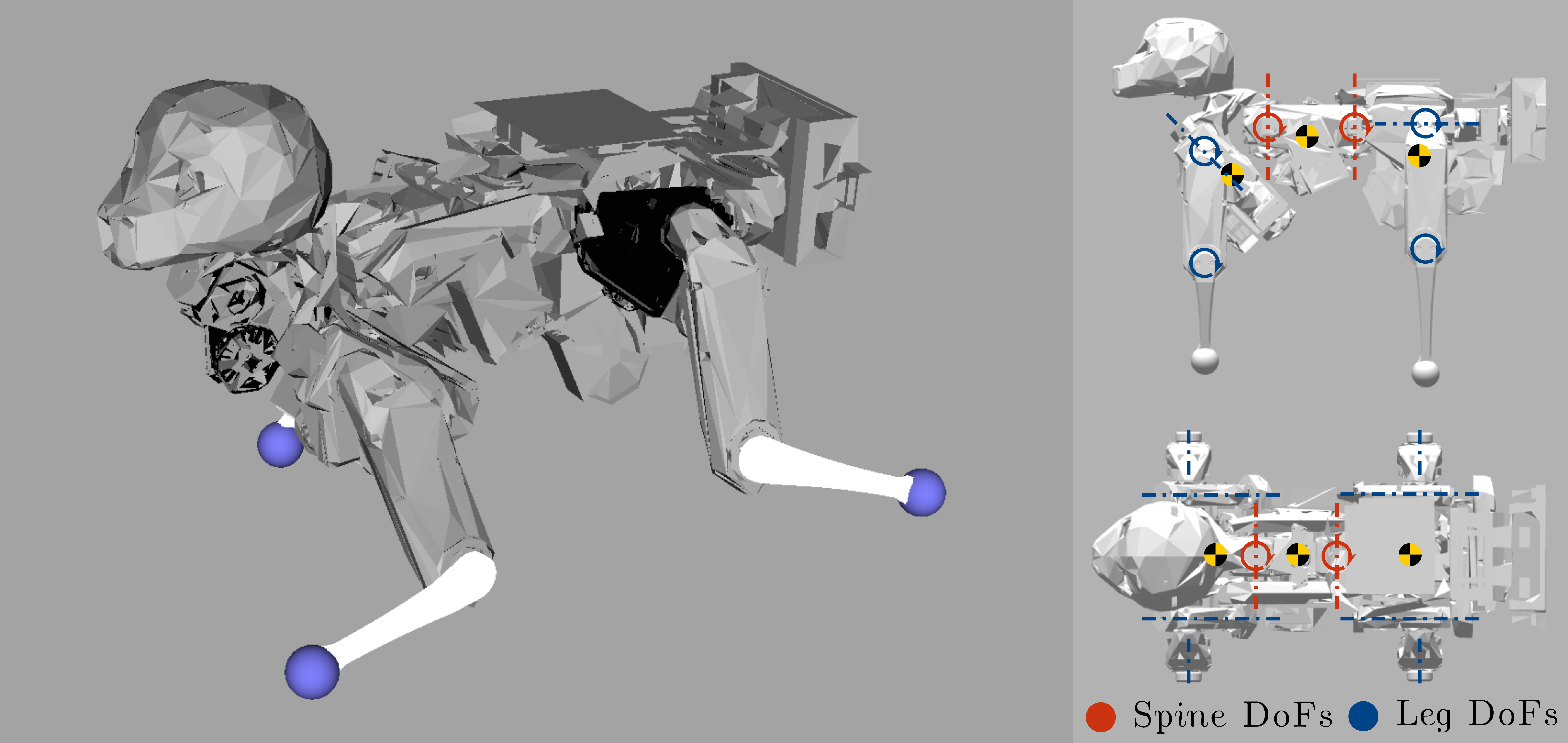}
    \caption{Left: the full robot being simulated. The head is purely aesthetic and given zero physical properties. Right: Joint axes and rigid body COMs}
    \label{fig:full_simulated_robot}
\end{figure}

The more complex robot used in the dart simulation is shown in Figure~\ref{fig:full_simulated_robot}. It attempts to model the approximate size and shape of a toy poodle.

We use the DART simulator \cite{dart} to perform simulations with the theoretical robot model. Our simulation environment is a single plane with default friction and restitution coefficients. The simulation is stepped with a 2.5ms time-step, resulting in a simulated control frequency of 400 Hz. The robot is loaded from a URDF file available alongside the simulation code on GitHub (Appendix~\ref{apx:accompanying_code}). Of note, the actuator parameters in Table~\ref{tab:actuator_parameters} are determined from the same 270Kv brushless DC motors used by \cite{Solo8_2020} and \cite{Solo12_2021}, with either 9:1 or 16:1 reduction ratios. We do not consider motor efficiency when calculating maximum applicable torque and velocity.

\begin{table}[]
    \centering
    \begin{tabular}{c|c|c|c|c}
         Joint & $\omega_{max}$ (rads\textsuperscript{-1}) & $\tau_{max}$ (N) & $\theta_{min}$ (rad) & $\theta_{max}$ (rad) \\
         Spine Pitch & 46.875 & 4.32 & $-\frac{\pi}{12}$ & $\frac{\pi}{12}$ \\
         Spine Yaw & 83.333 & 2.43 & $-\frac{\pi}{12}$ & $\frac{\pi}{12}$ \\
         Hip Roll (left) & 83.333 & 2.43 & $-\frac{\pi}{18}$ & $\frac{\pi}{4}$ \\
         Hip Roll (right) & 83.333 & 2.43 & $-\frac{\pi}{4}$ & $\frac{\pi}{18}$ \\
         Hip Pitch & 46.875 & 4.32 & $-\frac{3\pi}{4}$ & $\frac{3\pi}{4}$ \\
         Knee Pitch & 46.875 & 4.32 & $-\frac{4\pi}{5}$ & $\frac{4\pi}{5}$
    \end{tabular}
    \caption{Parameters used to model the robot actuators. $\omega_{max}$: Maximum angular velocity, $\tau_{max}$: Maximum torque, $\theta_{min}$: Joint minimum position, $\theta_{max}$: Joint maximum position}
    \label{tab:actuator_parameters}
\end{table}

\section{Notation}
\label{sec:notation}

We create two sets to simplify mathematical notation, $\mathfrak{s}$ contains the indices of all spine joints, and  $\mathfrak{f}$ contains the indices of the four feet. Foot positions and forces are represented by $\mathbf{p}_i$ and $\mathbf{f}_i$ where $i\in\mathfrak{f}$ respectively. Joint angles, velocities and torques are $\theta_i$, $\dot{\theta}_i$, and $\tau_i$, where $i$ is the index in the state vector (see Section~\ref{sec:control}. Letters represent conventional English names for feet e.g. $fl$ meaning `front left'. For the spine joints they represent the facing and axis of rotation, e.g. $ry$ corresponds to `rear pitch' (rotation about the joint `y' axis). For rigid bodies they represent `front', `middle' and `back'.

\begin{align}
    \mathfrak{s}&\in(fy, fz, ry, rz) \\
    \mathfrak{f}&\in(fl, fr, rl, rr)
\end{align}

We make use of the Special Orthogonal Lie Group ($SO(3)$) and associated Lie Algebra ($\mathfrak{so}3$) in our optimisation routines, robot dynamics models, controller, and scoring systems. For clarity, a brief, non-rigorous overview of the relevant sections of Lie Groups is provided. The SO(3) group is the set of rotation matrices with positive-defined determinants. Each member may be constructed by taking the matrix exponent of a combination of the three generator matrices $\mathcal{G}_n$.

\begin{equation}
    \label{eqn:lie_generators}
    \mathbf{R} \in SO(3) = exp(\alpha \mathcal{G}_0 + \beta \mathcal{G}_1 + \gamma {\mathcal{G}_2})
\end{equation}

\begin{equation}
    \mathcal{G}_n \in \Biggl\{
    \begin{pmatrix}
        0 & 0 &  0 \\
        0 & 0 & -1 \\
        0 & 1 &  0 \\
    \end{pmatrix},
    \begin{pmatrix}
        0 & 0 & 1 \\
        0 & 0 & 0 \\
       -1 & 0 & 0 \\
    \end{pmatrix},
    \begin{pmatrix}
        0 & -1 & 0 \\
        1 &  0 & 0 \\
        0 &  0 & 0 \\
    \end{pmatrix}
    \Biggr\}
\end{equation}

This lets us define the vector representation for the associated Lie Algebra. 

\begin{equation}
    \label{eqn:lie_algebra}
    \mathbf{r} \in \mathfrak{so}(3) = (\alpha, \beta, \gamma)^T
\end{equation}

The sum of generators is equivalent to a skew symmetric matrix constructed from the vector representation of the Lie Algebra. This allows us to define two maps, hat and vee, which map from the vector representation to the matrix representation and vice versa. In a slight abuse of notation, we use a hat diacritic above a variable to represent the \textit{matrix exponential of} the hat map (Equation~\ref{eqn:hat_notation}). Likewise a vee diacritic under a variable to represent the vee map of the matrix logarithm (Equation~\ref{eqn:vee_notation}).

\begin{equation}
    \label{eqn:hat_notation}
    \widehat{\mathbf{x}} \xrightarrow{}\bf{X}\ \ \ \ (\mathfrak{so}(3)\xrightarrow{} SO(3))
\end{equation}

\begin{equation}
    \label{eqn:vee_notation}
    \uwidevee{\mathbf{X}} \xrightarrow{} \bf{x}\ \ \ \ (SO(3)\xrightarrow{} \mathfrak{so}(3))
\end{equation}

Finally, we use the symbols $\mathbf{r}, \mathbf{R}, \mathbf{v}, \boldsymbol{\omega}$ to represent the position, orientation, linear velocity and angular velocity of the robot COM with respect to the global origin.

\section{Control System}
\label{sec:control}

\subsection{Control Overview}

\begin{figure}
    \centering
    \includegraphics[width=0.9\textwidth]{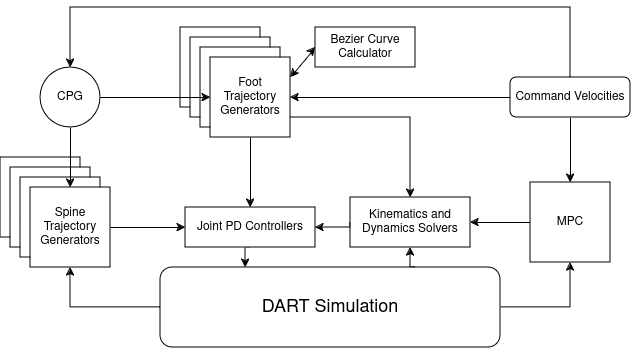}
    \caption{The full control system used to generate robot motion, with the CPG and MPC}
    \label{fig:control_system}
\end{figure}

The full system described in detail below can be seen in Figure~\ref{fig:control_system}. The central pattern generator (CPG) synchronises spine and leg trajectory generator cycles. The trajectory generators generate paths for their respective joint / leg to follow. MPC is used to determine the required leg ground-reaction forces (GRFs)  for stance-phase legs in real-time, with Jacobian-transpose inverse dynamics (ID) use to compute joint torques. PD controllers generate torques when joints are required to be following trajectories, feed-forward torque for the spine joints is calculated via ID and added to the commanded spine joint torque commands. The entire cycle is simulated to run at 200 Hz, despite the simulator time-step being 400 Hz, to verify real-time capabilities. The simulator feeds back the robot state to the MPC, dynamics solvers and trajectory generators as necessary.

The  two pitch ($\theta_{fy}, \theta_{ry}$) and two yaw ($\theta_{fz}, \theta_{rz}$) DoFs in the spine and 12 leg DoFs shown in Figure~\ref{fig:full_simulated_robot} are modelled as perfect hinge joints about their respective axis. 

\begin{figure}
    \centering
    \includegraphics[width=0.5\linewidth]{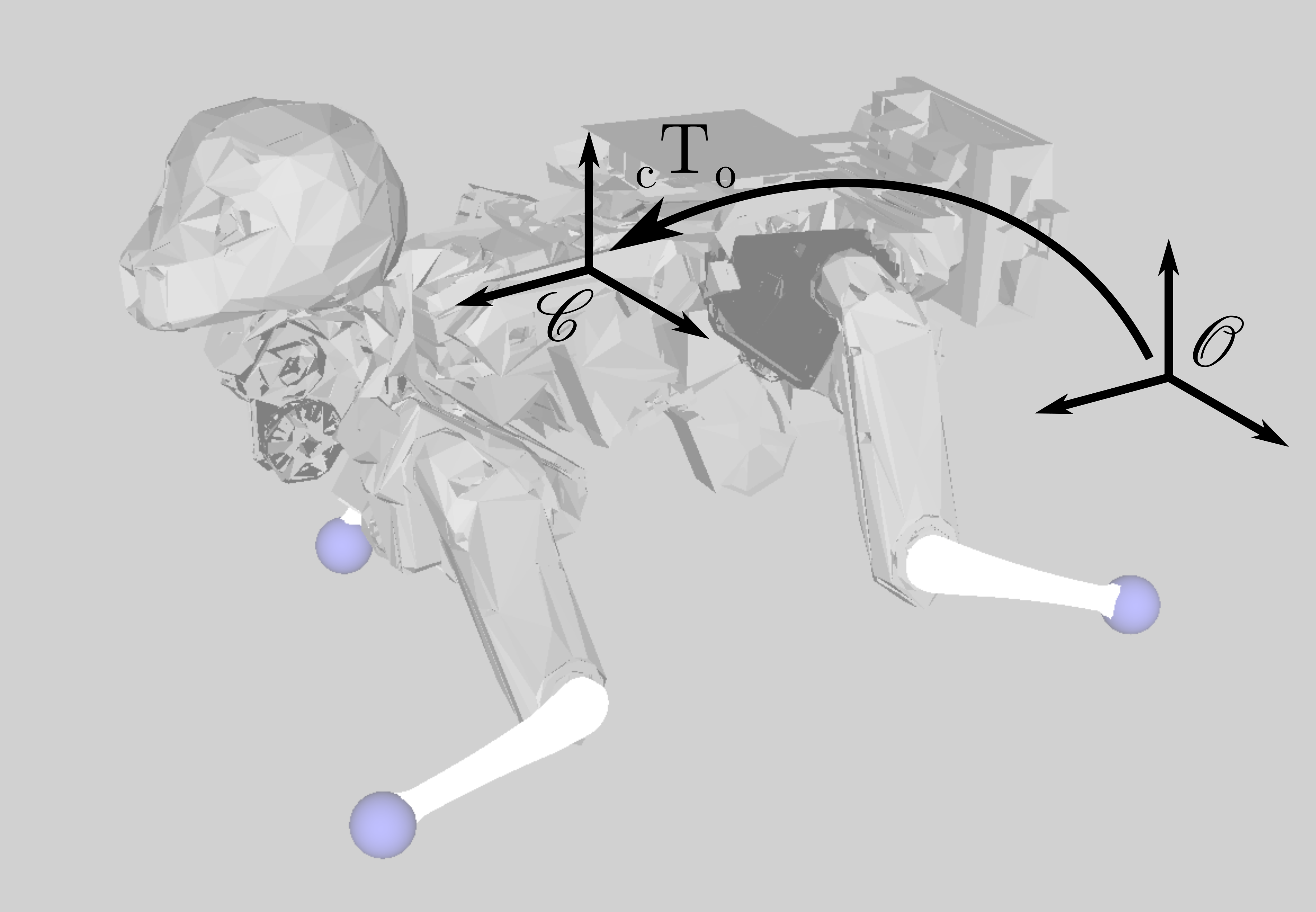}
    \caption{The world frame, $\mathcal{O}$, and COM frame, $\mathcal{C}$, are connected via a 6-DoF free joint within the simulation environment.}
    \label{fig:com_and_world_frames}
\end{figure}

Finally, we include a 6-DoF virtual free joint between the base of the robot $\mathcal{C}$ and the world origin $\mathcal{O}$, such that the robot is capable of motion in the simulated world. This is shown in Figure~\ref{fig:com_and_world_frames}, and gives the simulation a total of 22 DoFs.

\subsection{Robot Kinematics}

To calculate forward kinematics (FK) for any body $i$ on the robot, we simply take the transform of its parent relative to the origin, and apply the transform of itself relative to its parent, as per Equation~\ref{eqn:joint_fk}. The function $\nu(i)$ returns the index of the parent of the i\textsuperscript{th} body.
We calculate forward kinematics (FK) via Equation~\ref{eqn:joint_fk}, where ${}_{o}\mathbf{T}_{i}$ is the transform from the origin frame to the i\textsuperscript{th} body frame, ${}_{o}\mathbf{T}_{\nu(i)}$ is the transform from the origin frame to the parent frame of i, and ${}_{\nu(i)}\mathbf{T}_{i}$ is the transform from the parent frame of i to the i\textsuperscript{th} body frame.

\begin{equation}
    {}_{o}\mathbf{T}_{i} = {}_{o}\mathbf{T}_{\nu(i)} \cdot {}_{\nu(i)}\mathbf{T}_{i}
    \label{eqn:joint_fk}
\end{equation}

For inverse kinematics (IK), we do not consider the orientation of the feet, and thus the Cartesian space target has 12 variables. However, the joint space has 16 variables, resulting in a one-to-many solution map for the IK solver. This redundancy means that there may be an infinite number of solutions, or zero solutions (if the target is out of reach) when attempting to solve IK.

Rather than solving a redundant IK problem with one of many popular approaches \cite{whitney_1969, balestrino_1984, wolovich_1984, buss_2005, wang_1991, aristidou_2011}, we instead solve the legs analytically, and explore 4 new spine trajectory generation strategies, solving spine positions via FK.

Each leg of the proposed robot has 3 degrees of freedom and thus has a one-to-one mapping to Cartesian space, so can be solved analytically. However, this requires prior knowledge of each hip joint's position and orientation, which is a function of the spine degrees of freedom. The position of the foot with respect to the hip joint $\bm{s_{i}}, i\in\mathfrak{f}$ is found to be:

\begin{equation}
    \bm{s_{i}} = \bm{R_{0}v_{0}} + \bm{R_{1}R_{0}v_{1}} + \bm{R_{2}R_{1}R_{0}v_{2}}
\end{equation}

Where $\bm{v_{i}}$ is the vector from the origin of the joint $i$ to the origin of the joint ${i+1}$ (to the foot in the case of the final joint), and $\bm{R_{i}}$ is its respective rotation matrix. We use IKFast \cite{ikfast} to generate an efficient IK solver for the leg in the hip coordinate frame.

The spine function of quadrupedal animals is generally agreed to be a combination of stride extension and energy minimisation \cite{hildebrand_1961, hoyt_1981, bertram_2008}, however little existing work has considered quadrupedal robots with 4 DoF active spines. Therefore in order to generate FK targets, we explore 4 new trajectory generation strategies for the 4 spine joints.

To solve inverse dynamics for the legs and spine, we use the Jacobian transpose approach, as per Equation~\ref{eqn:inverse_dynamics} where $\tau$ is the joint torques, $\mathcal{J}$ is the joint Jacobian, and $\mathbf{f}$ is the forces applied at the end effectors.

\begin{equation}
    \tau = \mathcal{J}^T \mathbf{f}
    \label{eqn:inverse_dynamics}
\end{equation}

\subsection{Central Pattern Generator}

A CPG generates sawtooth waves syncing leg motion and spine motion throughout a gait cycle. This varies from 0 to 1 at a rate depending on the total stance-swing duration of the foot trajectories:

\begin{equation}
    t_{cycle} = t_{stance} + t_{swing}
\end{equation}

Where $t_{cycle}$ is the entire period of the foot trajectory cycle, $t_{stance}$ is the duration of the stance (ground contact) section, and $t_{swing}$ the duration of time in the swing (air) section of trajectory. The saw tooth wave of the front left leg, $\phi$, is used as a reference point for other legs and the spine trajectory generators to position themselves against. Fixed phase offsets allow the legs to achieve consistent footfall patterns, and for the spine to position its peak and minimum range of motion at set points throughout the gait.

\subsection{Foot Trajectory Generator}

Foot swing trajectories are designed to follow pre-computed bezier curve trajectories \cite{hyun_2014}. A simple PD controller is used to determine torques to apply to the joints during the swing phase, which is experimentally deemed sufficient. During the stance phase, no trajectory is generated and PD controllers are disabled. Torque commands are calculated directly from the MPC output, and foot motion emerges from the system dynamics.

\begin{table}
    \centering
    \begin{tabular}{c|c|c|c|c|c|c}
         Gait & $t_{stance}$ & $t_{swing}$ & $\psi_{fl}$ & $\psi_{fr}$ & $\psi_{rl}$ & $\psi_{rr}$ \\
         Walk & 0.3 & 0.1 & 0 & 0.5 & 0.75 & 0.25 \\
         Trot & 0.2 & 0.1 & 0 & 0.5 & 0.5 & 0 \\
         Turn & 0.3 & 0.1 & 0 & 0.5 & 0.75 & 0.25 \\
    \end{tabular}
    \caption{Parameters used in swing and stance foot trajectory generation for walking, trotting and turning gaits.}
    \label{tab:swing_stance_parameters}
\end{table}

\begin{figure}
    \centering
    \includegraphics[width=0.9\linewidth]{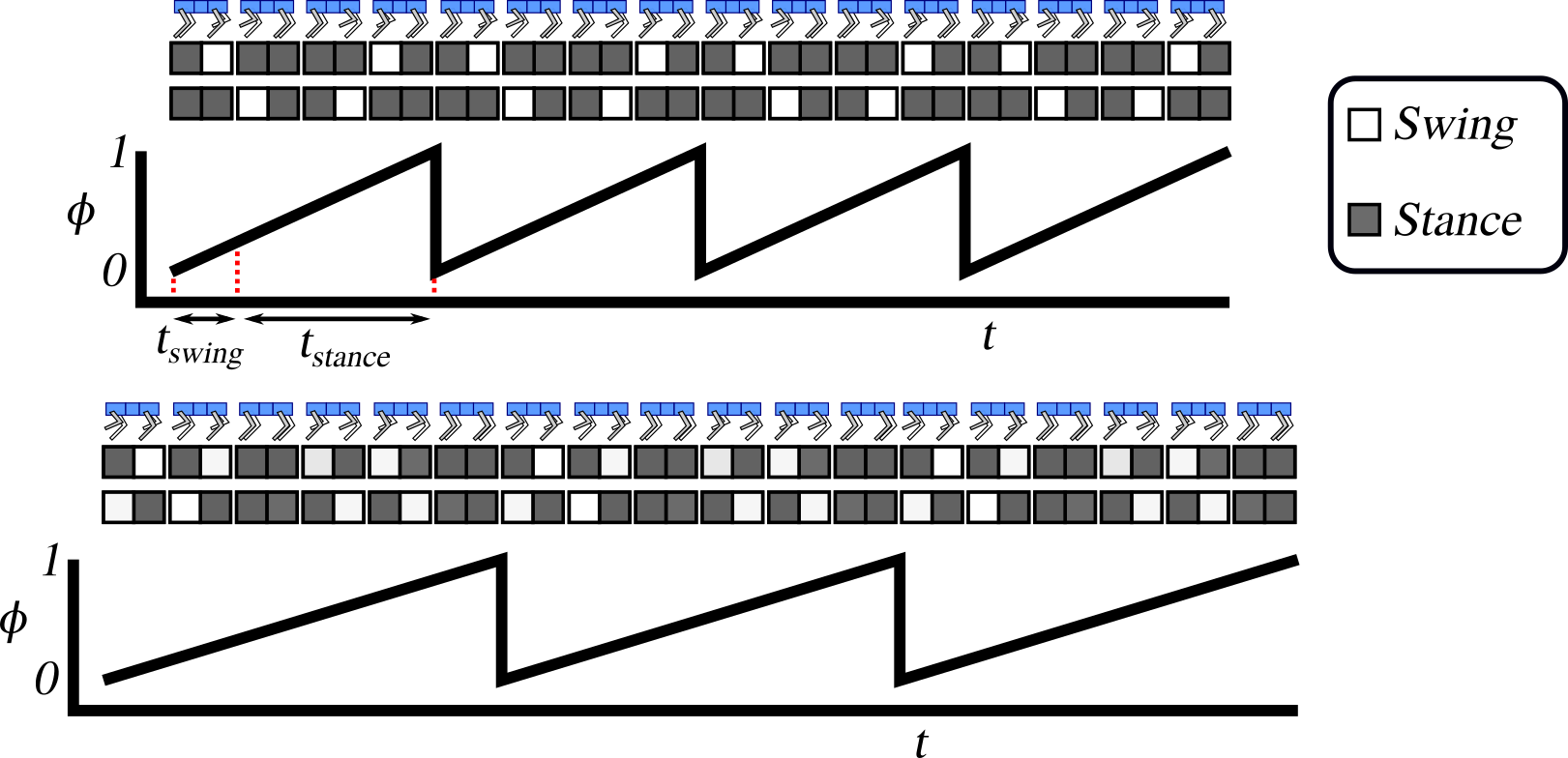}
    \caption{Footfall patterns for the walking (above) and trotting (middle) gaits. The sawtooth wave from the CPG is shown below each pattern.}
    \label{fig:footfall_patterns.png}
\end{figure}

Swing and stance phases controlled by the CPG have properties shown in Table~\ref{tab:swing_stance_parameters}. $\psi_{i\in\mathfrak{f}}$ is the offset of the leg phase variable relative to $\phi$. Parameters were chosen by hand to generate stable gait parameters.
The resulting footfall patterns in both walking and trotting gaits are shown along with the sawtooth wave from the CPG in Figure~\ref{fig:footfall_patterns.png}.

\subsection{Spine Control Strategies}
\label{sec:spine_control}

Prior research on active spines (see Section~\ref{sec:active_spine_robots}) uses direct user-control \cite{li_2023}, sinusoidal patterns \cite{hyun_2014, li_2023} or leg/stability-augmenting trajectories \cite{ishii_2009, horvat_2017, Kuehn_2018, chen_2022}, while passive (Section~\ref{sec:passive_spine_robots}) or hybrid (Section~\ref{sec:hybrid_spine_robots}) spines either take advantage of passive spring dynamics \cite{phan_2017, cao_2012, zhang_2013, nakatani_2009, eckert_2015, koutsoukis_2016}, or control spring properties directly \cite{sakai_2007, tsujita_2011, lu_2023}.
However none compare the efficacy of different methods.
Therefore, this paper proposes four different spine trajectory generation methods, with the express purpose of exploring both active and emergent trajectory generation strategies. All spine motors are controlled via PD-based torque control, with $K_p$ the proportional constant and $K_d$ the derivative constant.

The `stiffness' strategy (Equation.~\ref{eqn:stiffness_strategy}) periodically varies the motor stiffness constant, to give the effect of a spring with adjustable spring constant in the manner of \cite{sakai_2007, tsujita_2011, lu_2023}. $\theta_d$ is the desired joint position, $C_1$ controls the variation of stiffness amplitude, $C_2$ the offset with respect to $\phi$ and $C_3$ the mean stiffness. $C_4$ controls the damping applied to the joint.

\begin{equation}
    \label{eqn:stiffness_strategy}
    \begin{aligned}
        \theta_d &= 0 \\
        K_p &= C_1 sin(2 \pi \phi + C_2) + C_3 \\
        K_d &= C_4
    \end{aligned}
\end{equation}

The `foot-tracking' strategy (eqn.~\ref{eqn:foot_tracking_strategy}) poses that spine positions should complement the foot positions in the vein of \cite{ishii_2009, horvat_2017, Kuehn_2018, chen_2021}. This is done by extending or contracting based on the average foot distance from the neutral position, as well as rotating the hips to align with the relative foot angle from the neutral axis (y-axis). $C_1$ controls the range of motion, $C_2$ controls the stiffness and $C_3$ controls the damping. For all $i\in\mathfrak{f}$, $\mathbf{p}^i$ is the position of the foot relative to the chassis, while $\mathbf{c}^i$ is the neutral (centre) position of the foot stance phase in the same frame. An x subscript indicates the x component of the vector. $\hat{\mathbf{y}}$ is the unit y axis. $\theta_d^j$ for $j\in\mathfrak{s}$ is the target joint position.

\begin{equation}
    \label{eqn:foot_tracking_strategy}
    \begin{aligned}
        \theta^{fy}_d &= C_1 \frac{p_x^{fl} - c_x^{fl} + p_x^{fr} - c_x^{fr}}{2} \\
        \theta^{fz}_d &= C_1 arccos\biggl(\frac{\mathbf{p}^{fl} - \mathbf{p}^{fr}}{|\mathbf{p}^{fl} - \mathbf{p}^{fr}|}\cdot \mathbf{\hat{y}}\biggr) \\
        \theta^{ry}_d &= C_1 \frac{p_x^{rl} - c_x^{rl} + p_x^{rr} - c_x^{rr}}{2} \\
        \theta^{rz}_d &= C_1 arccos\biggl(\frac{\mathbf{p}^{rl} - \mathbf{p}^{rr}}{|\mathbf{p}^{rl} - \mathbf{p}^{rr}|}\cdot \mathbf{\hat{y}}\biggr) \\
        K_p &= C_2 \\
        K_d &= C_3
    \end{aligned}
\end{equation}

Finally, the `real-dog time-varying' strategy (Equation.~\ref{eqn:time_varying_strategy}) proposes a bi-periodic sin wave for the pitch degrees of freedom and mono-periodic sin wave for the yaw degrees of freedom (as seen in dogs by Wachs et al, 2016 \cite{wachs_2016}, and used in \cite{hyun_2014, li_2023}). Of the parameters, $C_1$ corresponds to the range of motion, $C_2$ the phase offset relative to $\phi$, $C_3$ the joint stiffness and $C_4$ the joint damping.

\begin{equation}
    \label{eqn:time_varying_strategy}
    \begin{aligned}
        \theta_{d} &= C_1 sin (4 \pi \phi + C_2) \\
        K_p &= C_3 \\
        K_d &= C_4
    \end{aligned}
\end{equation}

This strategy is also used with parameters from the optimisation search in Equation~\ref{eqn:spine_parameter_optimisation} rather than real-world values. In this case it is referred to as the `optimised time' strategy. Equations depend on various constants $C_1, ..., C_n$ and the phase parameter $\phi$. The offset from the phase in the time-varying strategies is chosen to align peak spine movement with the measured peaks and troughs in \cite{wachs_2016}.

Each joint has its own set of parameters defining the motion during the gait. These parameters were primarily found via a grid search attempting to minimise energy consumption, spine joint error (where applicable), body trajectory tracking error, gait instability, and in the case of stiffness and impedance also maximise spine joint range of motion. This cost function is shown in Equation~\ref{eqn:spine_parameter_optimisation}, with all averages calculated over the entire 10 second simulation.
$R^{winding}_i$ is the winding resistance of the simulated motors, 0.27 $\Omega$. Following the notation defined in Section~\ref{sec:notation}, $\mathbf{r}$, $\mathbf{R}$, $\mathbf{v}$ and $\boldsymbol{\omega}$ are the robot position, rotation, linear and angular velocities. $\theta_i$ is the i\textsuperscript{th} spine joint position. We use $\overline{\mathbf{x}}$ to calculate the average error of a vector variable $\mathbf{x}$, and $\sigma_i$ is the standard deviation in foot force applied by the i\textsuperscript{th} foot. $P$ is the average power applied by all spine joints. Superscript $d$ indicates a desired (target) value.
Finally, candidates with good scores were reviewed manually and chosen based on how natural they appeared to the researchers.

\begin{align}
    &\begin{aligned}
        \label{eqn:spine_parameter_optimisation}
        min\biggl( &W_1\overline{\mathbf{r}}
          + W_2\frac{\sum_{k=1}^{N}|\uwidevee{\mathbf{R}^k{\mathbf{R}^{d,k}}^T}|}{N}
          + W_3\overline{\mathbf{v}}
          + W_4\overline{\boldsymbol{\omega}} \\
          &+\sum_{i\in\mathfrak{f}} \Bigl[W_5\sigma_{i,fd} + W_6\sigma_{i,lo}
              + W_7\overline{\mathbf{f}_i} + W_8 max(\mathbf{f}_i)\Bigr] \\
          &+\sum_{i\in\mathfrak{s}}\Bigl[W_9\overline{\theta_i} + W_{10} (max(\theta_i) - min(\theta_i))\Bigr] \\
          &+W_{11}P\biggr)
    \end{aligned} \\
    &where \nonumber \\
    &\ \ \ \ \overline{\mathbf{x}} = \frac{\sum_{k=1}^{N}|\mathbf{x}^{d,k} - \mathbf{x}^k|}{N}, \mathbf{x} \in \mathbb{R}^{n \times 1} \nonumber \\
    &\ \ \ \ \sigma_i = \biggl|\sqrt{\frac{\sum_{k=1}^{N}(\mathbf{f}_i^k - avg(\mathbf{f}_i))^2}{N}}\biggr|, i\in\mathfrak{f} \nonumber \\
    &\ \ \ \ P = avg\biggl(\sum_{i\in\mathfrak{j}} \Bigl((\tau_i^k K_{v_i})^2 R^{winding}_i + \tau_i^k \dot{\theta}_i^k\Bigr)\biggr) \nonumber
\end{align}

Optimisation function weights are shown in Table~\ref{tab:spine_parameter_optimisation_constants} for each strategy. Note that the real-dog time-varying strategy is not optimised, the values are chosen manually. The parameters used in Equations~\ref{eqn:stiffness_strategy}-\ref{eqn:time_varying_strategy} vary between gaits,  and are available online. For further details see Appendix~\ref{apx:accompanying_code}.

\begin{table}
    \centering
    \begin{tabular}{c|c|c|c|c|c|c|c|c|c|c|c}
         Strategy & $W_1$ & $W_2$ & $W_3$ & $W_4$ & $W_5$ & $W_6$ & $W_7$ & $W_8$ & $W_9$ & $W_{10}$ & $W_{11}$ \\
         stiffness & 50 & 50 & 50 &  10 &  4 & 4 & 0.1 & $\frac{1}{800}$ & 0 & -2 & 0.002 \\
         foot-tracking  & 50 & 50 & 50 &  10 &  4 & 4 & 0.1 & $\frac{1}{800}$ & 10 & 0 & 0.002 \\
         time varying & 50 & 50 & 50 &  10 &  4 & 4 & 0.1 & $\frac{1}{800}$ & 10 & 0 & 0.002 
    \end{tabular}
    \caption{Parameters used in the cost function (Equation~\ref{eqn:spine_parameter_optimisation}) when searching for gait parameters. Weights were chosen to match the orders of magnitude of different components of the cost function.}
    \label{tab:spine_parameter_optimisation_constants}
\end{table}

\subsection{Model Predictive Control}
\label{sec:mpc}

With the increasing power available in modern computers, two primary control solutions for quadrupedal robots have emerged, \textit{optimisation-based} (such as convex optimisation \cite{gehring_2013}, whole body quadratic programming  \cite{risiglione_2022}, adaptive force control \cite{sombolestan_2021}, and MPC \cite{Camacho_Bordons_2007}) and \textit{learning-based} approaches \cite{aractingi_2023}. Since it is challenging to rapidly test the different spine control strategies outlined in Section~\ref{sec:spine_control} using learning based approaches, we choose to use an existing model predictive controller to control the ground reaction forces of the feet applied by our robot. By capturing the system dynamics, model-based methods are capable of performing with a variety of spine control strategies.

Recently, MPC approaches have been popular\cite{neunert_2018, bledt_2018, ding_2021, lecleach_2021, kang_2022}. Of note, Representation-Free Model Predictive Control (RF-MPC) as developed by Ding et al. used variation-based linearisation of the tangent-space to the rotation matrix to prevent the problem of gimbal lock experienced by Euler Angle-based approaches as Ding et. al \cite{ding_2021}. This allows the 18 variable state (Equation~\ref{eqn:rf-mpc-state}, the state of the chassis w.r.t the origin frame as per Figure~\ref{fig:com_and_world_frames}, using identical notation as \cite{ding_2021}) to be reduced to 12 variables when considering a small variation about an operating point at time-step $k$ (Equation~\ref{eqn:rf-mpc-state-linearised}), while maintaining high accuracy.

\begin{equation}
    X = (\mathbf{r}, \mathbf{R}, \mathbf{v}, \boldsymbol{\omega})^T
    \label{eqn:rf-mpc-state}
\end{equation}

\begin{equation}
    \mathbf{R}_k = \widehat{\delta \uwidevee{\mathbf{R}}_k}\mathbf{R}_0
    \label{eqn:rf-mpc-state-linearised}
\end{equation}

Following recent quadrupedal MPC literature, the legs of the robot are treated as weightless and thus their dynamic contributions are ignored. In addition, we assume that over a small time-period the spinal movements are small, and thus the robot may be approximated as a single rigid body, with forces applied at the foot positions. This allows us to control the required foot reaction forces with RF-MPC \cite{ding_2021}. No modifications are made to RF-MPC, as the spine movements are relatively small, the change to robot inertia and centre of mass location are not large enough to cause divergence from a high frequency real-time MPC controller. As in RF-MPC \cite{ding_2021}, qpSWIFT is used to solve the quadratic programming formulation of MPC in real-time.

\subsection{Joint Controllers}

We simulate real robot hardware compute requirements by only running the MPC every other simulation step, limiting it to a simulated 200 Hz update rate, below the 250+ Hz achieved on real hardware in \cite{ding_2021}. All joints are controlled via torque control, either with direct force output (leg joints in stance phase) or via PD torque controllers (all other joints). In order to compensate for non-negligible gravity and leg force effects, the spine joints additionally apply feed-forward torques calculated from inverse dynamics. During intermediate simulation steps the previously calculated joint torques are reapplied without additional computation.

\subsection{Experiment Setup}

To compare the various strategies of spine trajectory generation, we consider two main metrics. First we explore the physical effects of the spine upon the gaits, by comparing the CoT for each strategy.
CoT is calculated per Equation~\ref{eqn:cot}, with $P$ the average power consumption, $m$ the robot mass, $g$ the gravitational acceleration magnitude, and $v$ the velocity.

\begin{equation}
    CoT = \frac{P}{mgv}
    \label{eqn:cot}
\end{equation}

Secondly, we consider the effect of the different strategies upon participant perception of how natural the motion appears. We compare against a baseline fixed-spine strategy, where all 4 spinal degrees of freedom are locked at their zero positions, and no power is applied to them (since, a fixed-spine robot would not have joints in those positions). The simulated fixed-spine robot still contains the weight of the spine motors, therefore using identical mass and inertia matrices. It is noted that a robot without a spine would weigh less, and therefore use less energy, however this makes accurate comparisons challenging, as other effects such as inertial effects would change the dynamic properties of the simulated systems.

The robot is tested across a range of velocities, with walking, trotting, and turning gaits. Walking and trotting gaits are tested with purely linear velocities in the x-axis, with overlap between both gaits in order to find the CoT at various velocities and the natural gait transition velocity. The turning gait uses a fixed walking velocity of 0.3 ms\textsuperscript{-1} while varying the angular velocity.

\subsection{Subjective Experiment}

In order to estimate the naturalness of the various strategies, a subjective experiment is conducted by showing participants videos of the gaits and asking them to compare the naturalness between the videos.

For each spine trajectory strategy, the robot is recorded walking in a straight line or curved line, or trotting in a straight line for 10 seconds. The commanded velocities are given in Table~\ref{tab:velocity_commands}. The robot starts in a standing position, and is commanded to accelerate at 0.5 ms\textsuperscript{-2} and -1.0 rads\textsuperscript{-2} until it reaches the commanded velocities.

\begin{table}
    \centering
    \begin{tabular}{c|c|c}
         Gait & Linear Velocity (x-axis, ms\textsuperscript{-1}) & Angular Velocity (z-axis, rads\textsuperscript{-1})\\
         Walking & 0.3 & 0\\
         Turning & 0.3 & -0.5\\
         Trotting & 0.6 & 0
    \end{tabular}
    \caption{Command velocities given for the 4 different gaits used in experiments. Values were chosen empirically to give gaits with sufficient velocity and gait clarity.}
    \label{tab:velocity_commands}
\end{table}

We note that due to the different resulting CoTs of the strategies as shown in Figure~\ref{fig:cot_plot}, Section~\ref{sec:sim_res}, it is challenging to fairly choose perfect parameters for the videos. Our logic for choosing the values outlined in Table~\ref{tab:velocity_commands} are as follows:
\begin{itemize}
    \item All strategies performed well in walking at 0.3 ms\textsuperscript{-1}, even though some may wish to transition to trotting around this value.
    \item At 0.6 ms\textsuperscript{-1}, all strategies performed more efficiently in a trotting gait than a walking gait, while maintaining stability.
    \item The range of stable angular velocities for all gaits was 0.0-1.0 rads\textsuperscript{-1}, and thus the centre point was used for the turning experiment.
\end{itemize}

\begin{figure}
    \centering
    \includegraphics[width=0.75\textwidth]{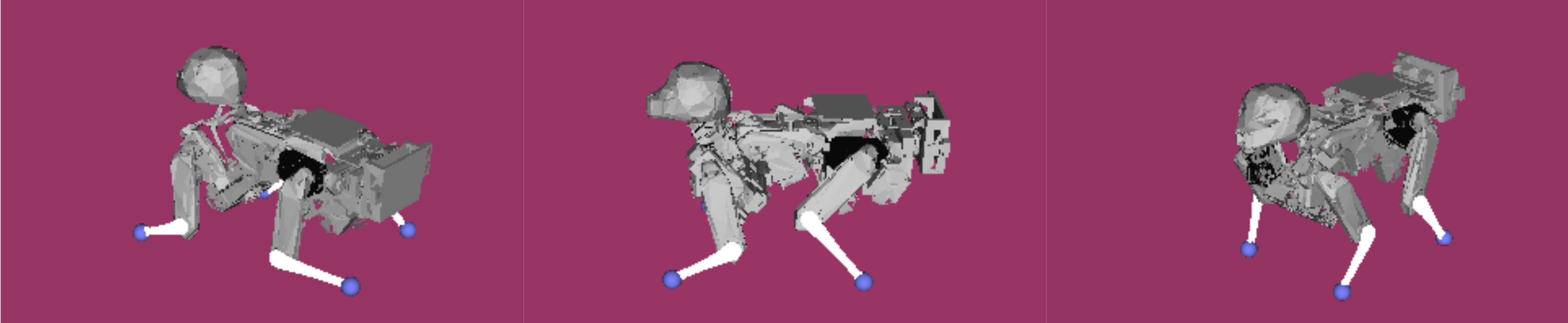}
    \caption{Perspective view of the robot in simulation. The robot travels from right to left. In this case it is turning as well.}
    \label{fig:tracking_perspective}
\end{figure}

All gaits are recorded by a 3D perspective camera in a fixed location, which tracks the centre of the robot chassis. This allows experiment participants to see the robot from several different angles, but ensures that all strategies are compared from identical angles. The camera height is chosen to give the same approximate perspective people have when looking at small dogs in real life.
An example of the perspective is given in Figure~\ref{fig:tracking_perspective}.

\begin{figure}
    \centering
    \includegraphics[width=0.9\textwidth]{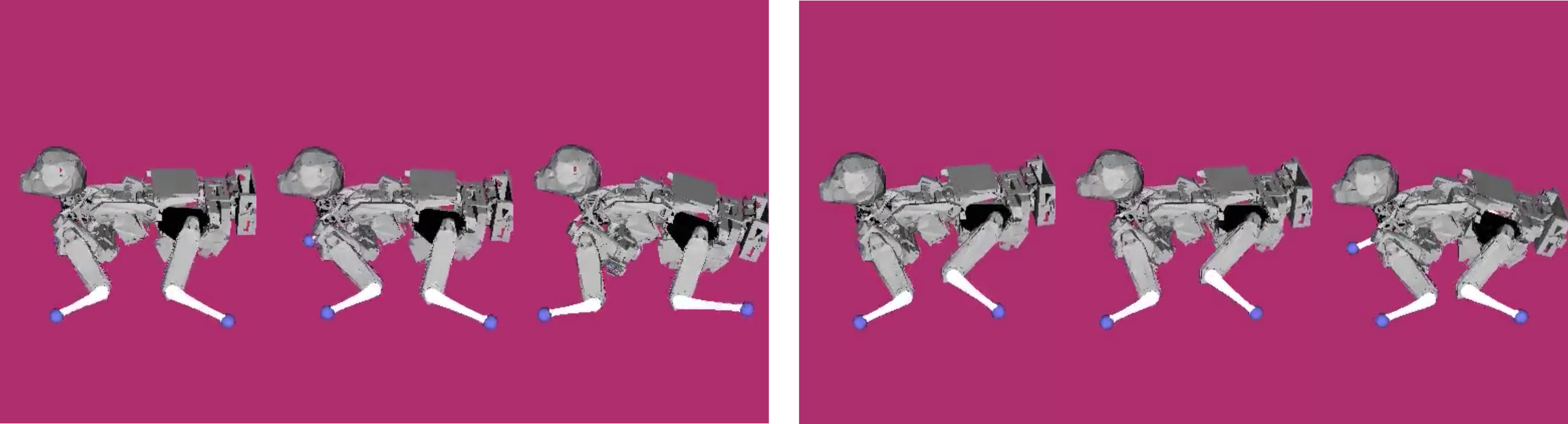}
    \caption{Two frames of a video participants were asked to watch, side-by-side. The three robots are walking right to left with different spine trajectory generators, and the frames are also in right-to-left order. The full videos are available online. See Appendix~\ref{apx:accompanying_code}.}
    \label{fig:spliced_video_frame}
\end{figure}

It is challenging to fairly rate the naturalness of a single gait or video in isolation, as there is no reference point for the audience to compare against. Additionally, comparing all 5 gaits next to each other is very challenging. As such, each strategy is compared with one other strategy and the fixed-spine baseline gait, by splicing all three videos together side-by-side, as shown in Figure~\ref{fig:spliced_video_frame}. This is repeated for every combination of trajectory generation strategy, ensuring that every gait is compared with every other gait an equal number of times, and the same number of times in each perspective and in each gait. Comparisons are always made with the spliced video contents in the same order, but the order in which videos are shown is randomised.

Participants were asked to select which of the three gaits shown in a video is the most natural, which is the least natural, and which one would they most be inclined to interact with. We hypothesise the following:

\begin{enumerate}
    \item A high positive correlation with naturalness and inclination to interact
    \item A positive correlation with energy efficiency and naturalness
    \item The real-dog time dependant gait to be rated as the most natural gait
    \item The fixed-spine baseline gait to be rated as the least natural gait
\end{enumerate}

The reason for hypothesis \#3 is that it is the only strategy based upon real-world scan data of healthy dogs. Hypothesis \#4 is proposed because real dogs do not walk with rigid spines, this is a definitely `robotic' or insectoid walking pattern.

In addition to the comparisons, an optional question was given for each video asking participants why they voted the particular way they did. This helps future work in its direction, or explain any strange results.

\begin{figure}
    \centering
    \includegraphics[width=0.8\linewidth]{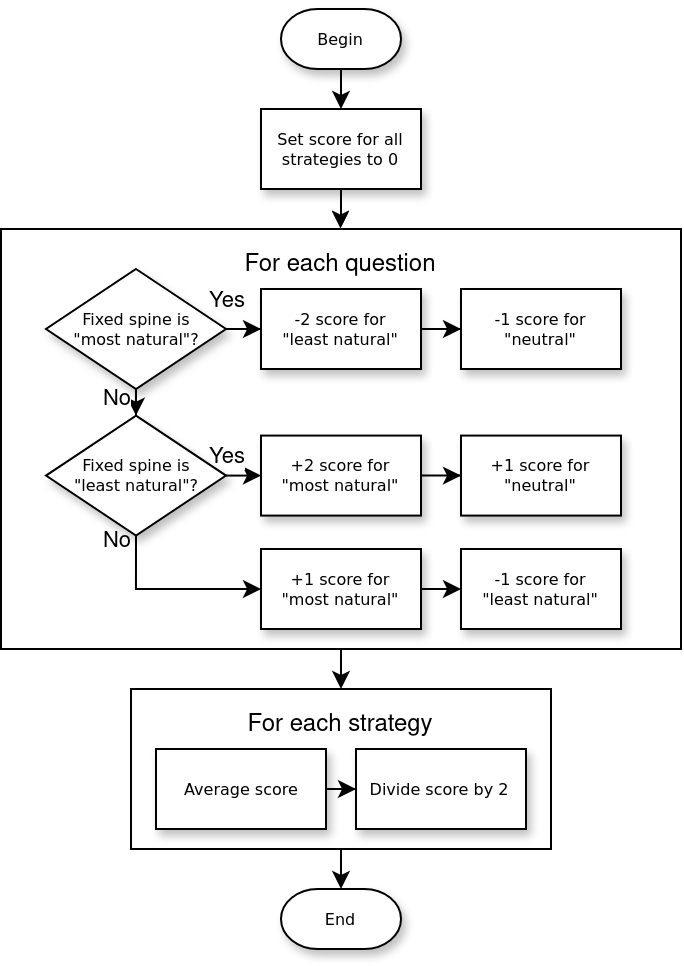}
    \caption{Algorithm to score strategy naturalness relative to a fixed spine baseline.}
    \label{fig:scoring_metric}
\end{figure}

In order to score the gaits, we construct a simple metric scoring relative to the baseline, and normalising by the total number of votes in the data being compared, shown in Figure~\ref{fig:scoring_metric}.
This gives each strategy a score ranging from -1 to 1, where a positive score is more natural than the baseline. The total score is then the mean score across all candidates.

\section{Results}

\subsection{Simulation Results}
\label{sec:sim_res}

\begin{figure}
    \centering
    \includegraphics[width=0.9\linewidth]{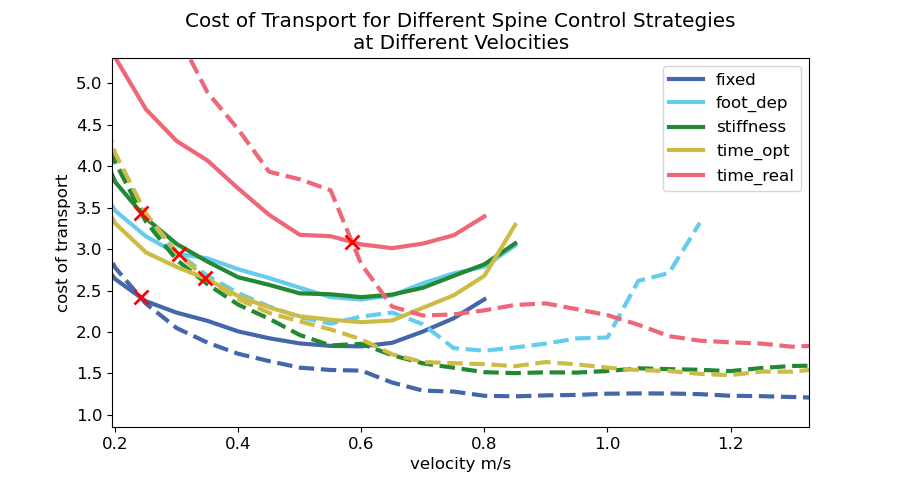}
    \caption{CoT for all strategies across a range of velocities. The leftmost curves correspond to the walking gaits, while the rightmost curves are the trotting gaits. Solid lines indicate walking gaits, dashed lines indicate trotting gaits, and} red crosses mark the walk-trot CoT intersection points.
    \label{fig:cot_plot}
\end{figure}

The calculated CoT across a range of velocities for both walking and trotting gaits is shown in Figure~\ref{fig:cot_plot}. At velocities higher than 0.8 ms\textsuperscript{-1} the walking gait is unstable for all strategies; trotting gaits are stable beyond 1.5 ms\textsuperscript{-1} for all strategies except foot-tracking, which is only stable to 1.2 ms \textsuperscript{-1}. We can see that relative CoT ranking does not change significantly with velocity, though the stiffness strategy performs better in the trotting gait than the walking gait, relative to the other strategies. At all velocities, the fixed spine gait outperforms the spine enabled gaits, as the four spine joints draw no power at all.

The best performing spine strategy for walking is the time-optimised strategy. While trotting the time-optimised strategy ties with the stiffness design. This suggests that at lower speeds, the flexibility of active control is more beneficial, while at higher speeds harmonic oscillations become more dominant.

\begin{figure}
    \centering
    \includegraphics[width=0.9\linewidth]{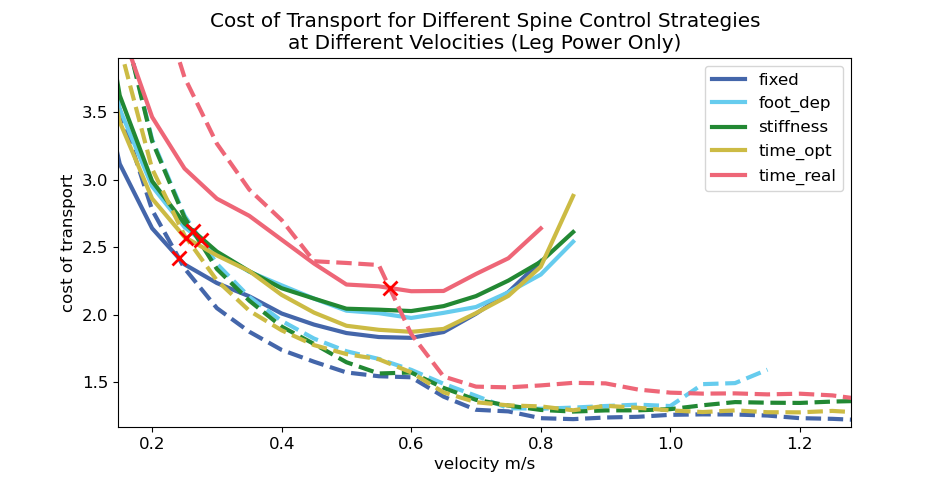}
    \caption{CoT for all strategies, ignoring the cost of actuating spine joints. The leftmost curves correspond to the walking gaits, while the rightmost curves are the trotting gaits. Solid lines indicate walking gaits, dashed lines indicate trotting gaits, and} red crosses mark the walk-trot CoT intersection points.
    \label{fig:cot_legs_only}
\end{figure}

When ignoring the power drawn by the spine motors as per Figure~\ref{fig:cot_legs_only}, the same pattern emerges, but the differences are much smaller. Once again, the real-dog gait has the highest CoT, while the fixed gait still outperforms spine-enabled gaits. This is contrary to the results in other works, which found that actuating the spine in a manner similar to our proposed foot-tracking strategy reduced the power drawn by the legs \cite{Kuehn_2018}. Our results also suggest that even using purely passive spine components would not offer any improvements over a fixed spine from an energy efficiency perspective, as the spine movements result in higher leg power usage.

The natural gait transition point in quadrupedal locomotion occurs when trotting becomes more energy efficient than walking. With the exception of the real-dog strategy, this transition point between walking and trotting occurs at similar velocity for all strategies, marked as red crosses in Figure~\ref{fig:cot_plot} with raw values shown in Table~\ref{tab:gait_transitions}. Ignoring spine cost (Figure~\ref{fig:cot_legs_only}) slightly shifts the transition point for the foot-tracking and time-optimised strategies, but otherwise has minimal effect on the gait transition points.

\begin{table}[]
    \centering
    \begin{tabular}{c|c|c|c|c}
         Fixed Spine & Stiffness & Foot-Tracking & Time-Optimised & Real Dog \\
         0.2415 &  0.2425 & 0.3038 & 0.3463 & 0.5864
    \end{tabular}
    \caption{Gait transition velocity for no-spine and the four strategies.}
    \label{tab:gait_transitions}
\end{table}

\begin{figure}
    \centering
    \includegraphics[width=0.95\linewidth]{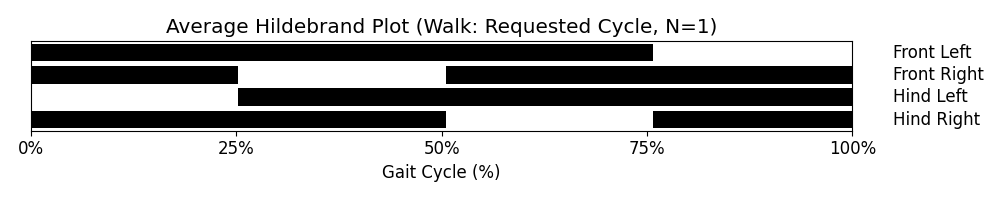}
    \includegraphics[width=0.95\linewidth]{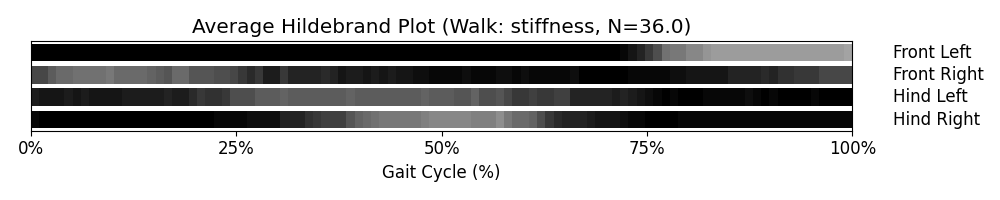}
    \includegraphics[width=0.95\linewidth]{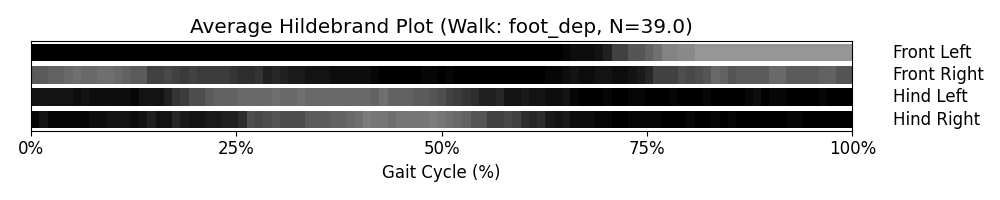}
    \includegraphics[width=0.95\linewidth]{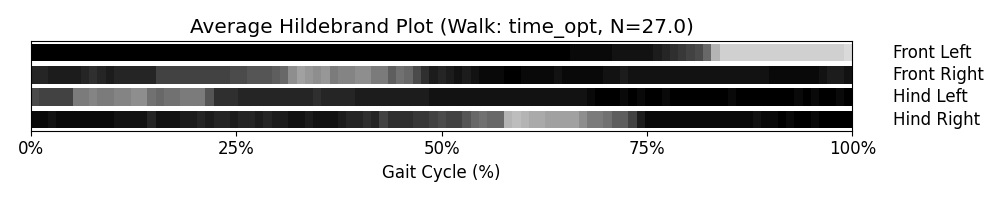}
    \includegraphics[width=0.95\linewidth]{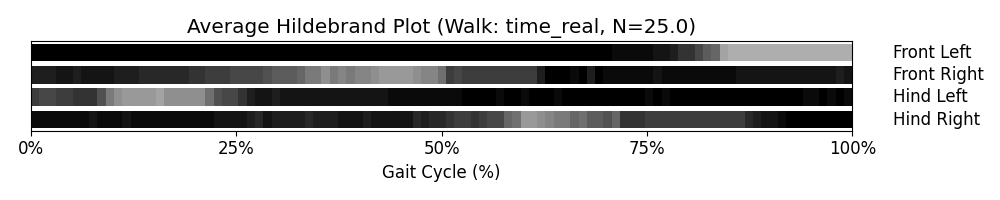}
    \includegraphics[width=0.95\linewidth]{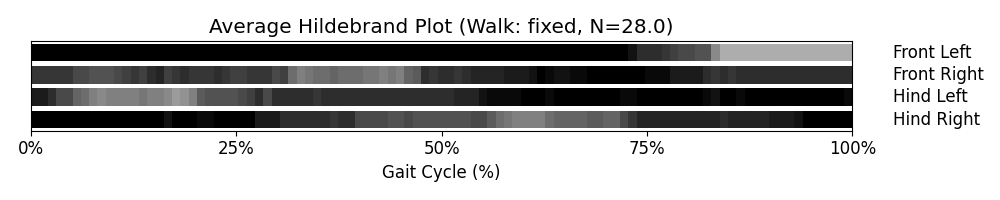}
    \caption{Hildebrand plots of walking gaits at 0.3 ms\textsuperscript{-1}, compared to the commanded cycle. Average stance over N cycles is shown, with 1 (stance) as pure black, and 0 (swing) as pure white}.
    \label{fig:walking_hildebrand}
\end{figure}

\begin{figure}
    \centering
    \includegraphics[width=0.95\linewidth]{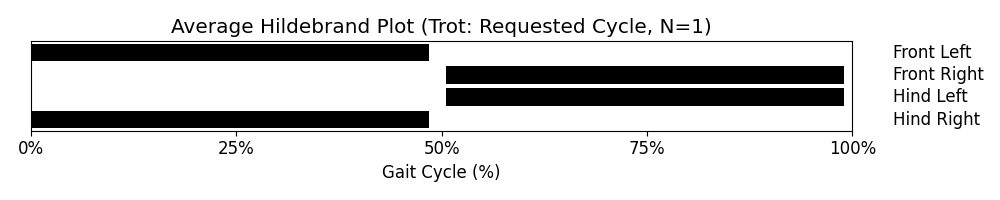}
    \includegraphics[width=0.95\linewidth]{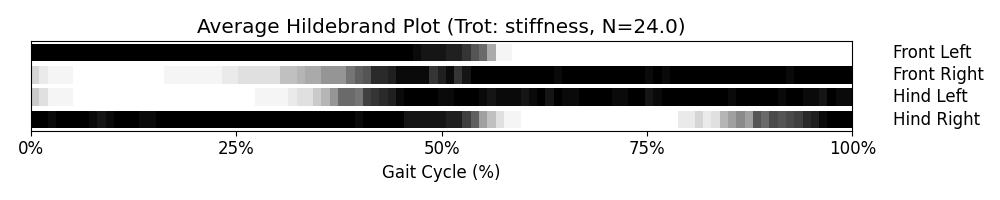}
    \includegraphics[width=0.95\linewidth]{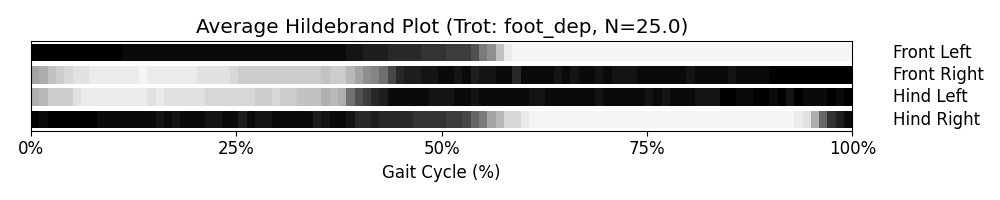}
    \includegraphics[width=0.95\linewidth]{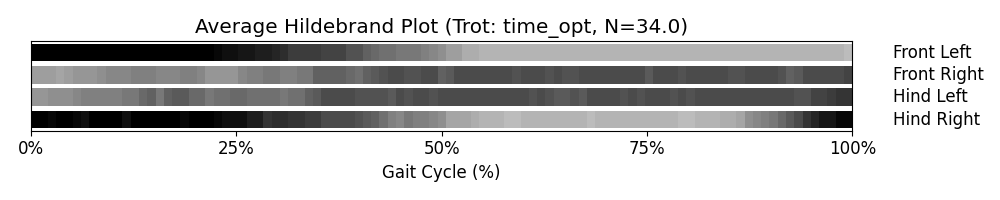}
    \includegraphics[width=0.95\linewidth]{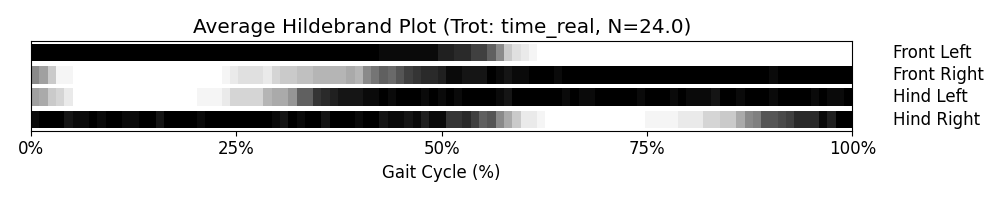}
    \includegraphics[width=0.95\linewidth]{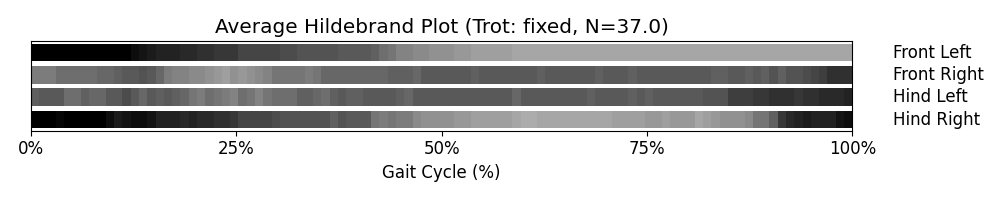}
    \caption{Hildebrand plots of trotting gaits at 0.6 ms\textsuperscript{-1}, compared to the commanded cycle. Average stance over N cycles is shown, with 1 (stance) as pure black, and 0 (swing) as pure white}.
    \label{fig:trotting_hildebrand}
\end{figure}

We additionally provide an ethological analysis of the strategies via Hildebrand plots \cite{hildebrand_1961, hildebrand_1968} in Figures~\ref{fig:walking_hildebrand} and \ref{fig:trotting_hildebrand}.
In walking (Figure~\ref{fig:walking_hildebrand}), foot-tracking and stiffness strategies do not track the reference cycle as well as other strategies, suggesting that these approaches introduce additional dynamic complexity leading to lag or noise. The noise in all approaches suggests somewhat unstable gaits (even with the fixed spine), which may be due to the relatively low frequency of leg-control, matching the 200 Hz MPC frequency.
Meanwhile in trotting (Figure~\ref{fig:trotting_hildebrand}), stiffness, foot-tracking and real-dog strategies reliably follow the requested footfall pattern. This suggests that appropriately chosen spine motion has a positive effect on gait stability at higher velocities.

\subsection{Subjective Results}

50 independent persons were asked to view the 18 generated videos on Amazon Mechanical Turks. One participant was dropped for regularly voting a gait to be the most and also least natural in a particular question, giving 49 total results. The questionnaire took an average of 15 minutes per person, and the participants were paid \$5 (USD) each. 

\begin{figure}
    \centering
    \includegraphics[width=0.7\textwidth]{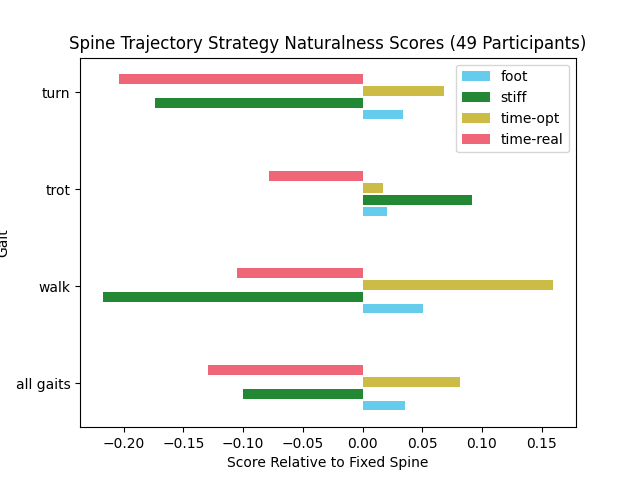}
    \caption{Subjective results across 49 participants. The baseline fixed-spine category has the number of votes halved, as it shows up twice as often as other categories.}
    \label{fig:all_participants_results}
\end{figure}

The overall results shown in Figure~\ref{fig:all_participants_results} show that the optimised time trajectory was deemed to be the most natural. The second most natural was the foot-tracking gait. The remaining two strategies performed worse than the fixed-spine strategy. Interestingly the score for the stiffness strategy is much higher in the trotting gait, moving from 4\textsuperscript{th} place to 1\textsuperscript{st} in this gait alone. Comments from the participants noted that it `didn't have too much or too little wiggle', or `moves its body with its legs'. Comments regularly suggested that excessive movement gave `actuated or robotic feelings'. This aligns with the low scores of the real-dog time varying strategy, which had both the highest movements and the lowest scores. However, some participants said that the larger movements made the gaits look \textit{less} robotic, and more `lively'.

\section{Discussion}

\subsection{Hypothesis \#1}
Hypothesis \#1 proposed that there would be a positive relation between naturalness and inclination to want to interact with the robot over other robots. In our results, `most natural' and `most likely to interact with' votes matched 862 times, and didn't match 20 times. This means that 97.7\% of votes matched between the two.

\subsection{Hypothesis \#2}
Here we expected a positive correlation with energy efficiency and naturalness. The naive results here are uncertain, as the most energy efficient strategy (fixed-spine baseline) was only the third-most natural - expected, given that it has four fewer active spine DoFs. Interestingly, the fixed-spine gait has the lowest CoT even when we discard the spine contributions (as per Figure~\ref{fig:cot_legs_only}). This implies that the current spine strategies are increasing the work required by the legs.

However, comparing only the spine-enabled strategies reveals that energy efficiency may be correlated with naturalness - as per Table~\ref{tab:relative_naturalness_and_energy_efficiency}, there is a clear relation in the trot and turning gaits, while the walking gait does not show much correlation.

\begin{table}[]
    \centering
    \begin{tabular}{c|c|c|c|c|c|c}
         Gait & N (Walk) & $\eta$ (Walk) & N (Trot) & $\eta$ (Trot) & N (Turn) & $\eta$ (Turn) \\
         foot & 2\textsuperscript{nd} & 1\textsuperscript{st} & 2\textsuperscript{nd}
              & 3\textsuperscript{rd} & 2\textsuperscript{nd} & 2\textsuperscript{nd} \\
         stiff & 4\textsuperscript{th} & 2\textsuperscript{nd} & 1\textsuperscript{st}
               & 1\textsuperscript{st} & 3\textsuperscript{rd} & 3\textsuperscript{rd} \\
         time-opt & 1\textsuperscript{st} & 3\textsuperscript{rd} & 3\textsuperscript{rd}
                  & 2\textsuperscript{nd} & 1\textsuperscript{st} & 1\textsuperscript{st} \\
         time-real & 3\textsuperscript{rd} & 4\textsuperscript{th} & 4\textsuperscript{rd}
                   & 4\textsuperscript{th} & 4\textsuperscript{th} & 4\textsuperscript{th}
    \end{tabular}
    \caption{Relative naturalness (N) and energy efficiency ($\eta$) ranking for the 4 gaits, ignoring the fixed-baseline.}
    \label{tab:relative_naturalness_and_energy_efficiency}
\end{table}

\subsection{Hypothesis \#3}
Hypothesis \#3 proposed that the real-dog time varying strategy would be the most natural. This gait was in fact the least natural according to the participants of this experiment. We think  this is due to the differences in size and weight distribution between the modelled robot and real animals data was taken from. Small animals and large animals have different walking gaits due to different physical constraints\cite{hildebrand_1968}, which applies to bio-inspired robots too. For example, consider a Toy Poodle quickly moving its feet compared to the long, slow strides of a Labrador. Our robot is closer in size to a Toy Poodle rather than a large dog, so Beagle data is used from \cite{wachs_2016}. Due to the virtual simulation environment, experimentees had no direct size reference, and may have made comparisons to larger dogs rather than small dogs.

\subsection{Hypothesis \#4}
The final hypothesis that the fixed-spine baseline strategy would be rated as the least natural strategy has also been proven false. The excessive movements of the real-dog time-varying strategy and the dragging motions performed by the stiffness strategy were perceived as even less natural.

\subsection{Ethological and Subjective Comparison}

We find it interesting to note that the most natural two trotting strategies have much more consistent footfall patterns than the fixed spine and time-optimised approaches. This may suggest a correlation between naturalness and gait consistency, however this pattern does not continue in the walking gait, and may be unrelated. Appropriate spinal motion has a positive effect on footfall consistency when trotting, and potentially also in higher velocity gaits.

This possible correlation between naturalness and consistency at higher velocities seems to align with the correlation between naturalness and energy efficiency, as shown in Table~\ref{tab:relative_naturalness_and_energy_efficiency}. Both are unclear in the walking case, with more closely matching results in trotting. To draw stronger conclusions, data from a larger range of velocities and other gaits would be required.

\section{Future Work}

None of the proposed spine control strategies were more energy efficient than the fixed spine alternative. Since existing work has shown that spines can make significant improvements to CoT \cite{hyun_2014}, further work could consider passive or hybrid spine systems. For example, average feed-forward torque in the spine joints could be provided by well-designed passive components. The stiffness strategy could be developed into a spring-based hybrid system rather than an active system that mimics an adjustable spring. The time-optimised strategy could benefit from similar improvements.
However, as we are also interested in the development of natural gaits with a focus on human interaction, we wish to have the capability and flexibility of active joint control in order to change to the most natural control strategy for a given gait. To improve our real-dog strategy, data from a larger variety of dog sizes should be analysed and compared across a range of robot sizes.

Additionally, our system does not consider the roll-DoF, due to conflicting opinions on its effect; this poses an exciting avenue for additional work. Rolling of the shoulders and hips may contribute to naturalness, by reducing the magnitude of pitch movements due to leg asymmetry. This may also reduce inertial motion, potentially increasing energy efficiency. We believe that reducing DoFs (1-pitch, 1-yaw, etc.) will increase energy efficiency at the cost of spine flexion and movement imitation capability. Thus designs must focus on the core purpose of the robot and how realistic motion needs to be.

Our current MPC system wraps around the spine trajectory generators, but does not model the internal rigid body dynamics of the moving spine. Another avenue for future research could consider modelling the internal spine dynamics within the MPC subsystem, or even controlling the spine joints directly using such a system. This may also improve the footfall consistency issues noted in Section~\ref{sec:sim_res}. Existing work models inertia changes due to variation in joint position, but does not consider the dynamic inertial effects of joint motion \cite{li_2023}.

Finally, we aim to develop the simulated robot into a hardware platform to perform more rigorous social experiments in the real world. Simulation results suggest that design improvements need to be made first should we wish to reduce energy consumption, but with naturalness of motion as a key focus this is less important.
A hardware platform would also enable research into facets of human-robot interaction (HRI) that are challenging to perform in simulation, such as imitating animals in touch-based interaction, or modelling physical interaction within the system controller.

\section{Conclusion}
In this work, we presented and evaluated 4 approaches to generating and controlling the spine trajectories within a spine-enabled quadrupedal robot. Strategies included foot-tracking similar to \cite{ishii_2009, horvat_2017, Kuehn_2018, chen_2021}, varying-stiffness in the vein of \cite{sakai_2007, tsujita_2011, lu_2023} and sinusoidal approaches based on real-data \cite{wachs_2016} similar to \cite{hyun_2014, li_2023}.
We compared the energy efficiencies of these approaches, finding that the CoT of sinusoidal and stiffness-based approaches were both lower than the foot-tracking approach across a range of velocities. However the fixed-spine baseline had a lower CoT than all approaches.
We note that prior work with spines achieves significantly lower CoT \cite{hyun_2014}, but makes no consideration of gait naturalness.

Simulation results suggest that improvements need to be made to the robot design before work can begin on a hardware implementation, due to large increases in energy draw at lower velocities. In particular, weight distribution and carefully positioned inertial bodies are very important when designing multi DoF spinal systems.


We found that some spine strategies improved the naturalness of motion, however excessive motion, or excessive slack had the opposite effect. While there was some relation between lower CoT and higher naturalness, the trend may not be particularly strong. We suggest that for HRI scenarios chasing naturalness the effects on energy efficiency may not be significant as higher scoring `natural' spine control strategies had similar CoTs, particularly at higher velocities. We also find that different control strategies have different perceived naturalness in different gaits, implying that HRI-focused robots should be able to switch strategy. This places limitations on potentially more energy efficient purely passive systems.

\section*{Ethical Approval Declarations}
This experiment was reviewed and approved by the ethical committee of our research institute (ethical review number 24-607). All the recruited subjects gave their formal consent following the proscribed procedure of the ethical committee.

\section*{Funding}
This work was partly supported by JST, Moonshot R\&D Grant Number JP-MJMS2011 (methodology conceptualization), and the Grant-in-Aid for Scientific Research on Innovative Areas JP22H04875 (evaluation experiments).

\section*{Author Biography}

Nicholas Hafner received his MEng in Mechanical Engineering at Imperial College London, and is now a Doctoral Candidate in the Department of Systems Innovation at Osaka University. His research interests include control theory, and the interaction between dynamic legged robots and humans.

Chaoran Liu obtained his Ph.D. from Osaka University Japan in 2015, and worked there as a specially appointed assistant professor. He joined ATR Hiroshi Ishiguro Laboratories in 2016, and the RIKEN Guardian Robot Project in 2021. He is currently with the National Institute of Informatics, from 2024. His main research interest is geometric machine learning and its applications in signal processing and interactive robots.

Carlos Toshinori Ishi received the Ph.D. degree in engineering from University of Tokyo, Japan, in 2001. He was with the JST/CREST Expressive Speech Processing Project from 2002 to 2004 at ATR Human Information Science Laboratories. He joined ATR Intelligent Robotics and Communication Laboratories in 2005, and became Group Leader of the Sound Environment Intelligence Research Group since 2013. He also joined the RIKEN Guardian Robot Project since 2020. His research interests include speech analysis, prosody, voice quality, paralinguistics, non-verbal expression, sound environment intelligence, dialogue robots and human-robot interaction.

Hiroshi Ishiguro received a D. Eng. in systems engineering from the Osaka University, Japan in 1991. He is currently Professor of Department of Systems Innovation in the Graduate School of Engineering Science at Osaka University (2009-) and Distinguished Professor of Osaka University (2017-). He is also visiting Director (2014) (group leader: 2002–2013) of Hiroshi Ishiguro Laboratories at the Advanced Telecommunications Research Institute and an ATR fellow. His research interests include sensor networks, interactive robotics, and android science. He received the Osaka Cultural Award in2011. In 2015, he received the Prize for Science and Technology (Research Category) by the Minister of Education, Culture, Sports, Science and Technology (MEXT). He was also awarded the Sheikh Mohammed Bin Rashid Al Maktoum Knowledge Award in Dubai in 2015. In 2020, he received Tateishi Prize. In 2021, he received an honorary degree of Aarfus University in Denmark.

\bibliographystyle{tfnlm}
\bibliography{main.bib}

\appendix

\section{Accompanying Data}
\label{apx:accompanying_code}
Code, videos, and raw data generated within this project is available to the public, hosted at: \url{https://github.com/Nickick-ICRS/ESTER}.

\end{document}